%% file: main_arxiv.tex
\documentclass{article}
\usepackage{iclr2027_conference,times}
\usepackage{amsmath,amssymb,amsfonts,bm}
\usepackage{graphicx}
\usepackage{booktabs}
\usepackage{multirow}
\usepackage{array}
\usepackage{xcolor}
\usepackage{url}
\usepackage[hidelinks]{hyperref}
\usepackage{microtype}
\usepackage{xspace}
\usepackage{enumitem}
\usepackage{placeins}
\usepackage{tikz}
\usetikzlibrary{arrows.meta,positioning,fit}
\hypersetup{
  pdftitle={VERA: Verifiable Feasibility Representations with Counterfactual Credit for Constrained Multi-Agent Control},
  pdfauthor={Bo Yin, Dongbo Li, Hongkai Chen, Jie Liu, and Guoliang Xing},
  pdfsubject={Preprint},
  pdfkeywords={multi-agent reinforcement learning, constrained control, feasibility representation, counterfactual credit, SAGIN}
}
\let\cite\citep
\newcommand{\R}{\mathbb{R}}

\newcommand{\chat}{\hat{\bm c}}
\newcommand{\cgt}{\bm c^{\star}}
\newcommand{\method}{VERA\xspace}

\title{\method: Verifiable Feasibility Representations with Counterfactual Credit\\for Constrained Multi-Agent Control}
\author{%
Bo Yin$^{1}$ \quad Dongbo Li$^{1,2,*}$ \quad Hongkai Chen$^{2}$ \quad
Jie Liu$^{1}$ \quad Guoliang Xing$^{2}$\\
$^{1}$Harbin Institute of Technology, Harbin, China\\
$^{2}$The Chinese University of Hong Kong, Hong Kong SAR, China\\[-0.2em]
\textnormal{\small $^{*}$Corresponding author: \href{mailto:ldb@hit.edu.cn}{ldb@hit.edu.cn}}\\[-0.2em]
\textnormal{\small Preprint, September 2026}}
\iclrfinalcopy
\begin{document}
\maketitle
\lhead{Preprint}

\begin{abstract}
Constrained multi-agent control requires more than predicting rewarding actions: an action can cease to be executable as contact windows, shared capacity, and deadlines change. We introduce \method, a centralized-training, decentralized-execution framework that separates feasibility estimation from credit assignment. Each actor predicts a five-dimensional \emph{verifiable feasibility representation} (VFR). After an action is proposed, exact action-conditioned margins available only during training supervise that representation, while a \emph{counterfactual group-relative advantage} (CGRA) ranks candidate representation--action pairs. Execution uses one actor pass and no privileged state. In a dynamic space--air--ground integrated network (SAGIN), \method obtains $55.33\pm3.60\%$ success with $0.45\pm0.81\%$ coverage violation, within $1.33$ points of a privileged-mask reference. With rewards matched over ten paired seeds, \method improves success over the strongest baseline by $8.74$ points ($p=0.023$) and reduces violation by $52.19$ points ($p=5.7\times10^{-8}$). A ten-seed $4\times2$ factorial attributes a $14.16$--$16.48$ point gain to CGRA across handcrafted, learned, random, and latent representations; evaluation on seven unseen topologies preserves a $24.33$--$30.02$ point advantage over multi-agent proximal policy optimization. From 10 to 40 users, success remains $50.1$--$53.8\%$, and VFR adds only $0.026$ ms to a central processing unit (CPU) actor step. Cross-domain tests further identify the governing condition: counterfactual credit succeeds when candidate scores respect shared constraints and fails under incompatible reward geometries. These results establish action-conditioned feasibility as an auditable training interface and counterfactual credit as a geometry-dependent optimization mechanism.
\end{abstract}

\input{sections/intro}
\input{sections/related}
\input{sections/method}
\input{sections/experiments}
\input{sections/conclusion}

\subsection*{Reproducibility statement}
The companion artifact includes the simulator, training and evaluation code, fixed seeds, run configurations, per-seed traces, checkpoints, frozen summaries, and independent recomputation tools. Appendix~\ref{app:details} documents the physical model, objectives, protocols, statistical tests, and artifact map. The ten-seed studies and 4,000-record VMAS audit are fully traceable; all 48 files in the trace archive pass SHA-256 verification.

\subsection*{AI use statement}
Generative AI tools assisted with language editing, manuscript organization, LaTeX, reference discovery, and figure-code refinement. They were not used to generate experimental data, execute experiments, select reported outcomes, or change statistical analyses. The authors verified the manuscript, citations, equations, figures, and numerical claims against primary sources and archived artifacts and take full responsibility for the submission.

\bibliographystyle{iclr2027_conference}
\bibliography{ref}

\appendix
\raggedbottom
\input{sections/appendix}
\end{document}

%% file: sections/intro.tex
\section{Introduction}
\label{sec:intro}
Multi-agent reinforcement learning (MARL) has become a practical route to decentralized control, with centralized critics, value factorization, and policy-gradient methods supporting increasingly large cooperative systems~\cite{07,08,20,47,49,revisiting_marl}. Yet most policies still compress two different questions into one action distribution: which action has produced high return, and which action remains physically executable now. In a dynamic physical system, the second question changes with contact duration, congestion, resource contention, and deadline slack. A policy can therefore exploit a stable reward correlation while failing to represent the mechanism that makes its decision feasible.

This distinction is especially sharp in a space--air--ground integrated network (SAGIN). Each ground agent must choose local execution, a low-Earth-orbit (LEO) satellite, or an unmanned aerial vehicle (UAV) edge node, together with an offloading ratio and bandwidth request. Satellite visibility, shared capacity, remote load, and end-to-end latency jointly determine whether the selected action can complete. Existing learning-based offloading and routing methods optimize return or expected constraints~\cite{09,21,22,23,24,25,28,29}, while a hard action mask can enforce feasibility only when exact instantaneous constraints are available at decision time. In our decentralized setting, those global quantities are unavailable to the deployed actor, although the simulator can compute them after an action is proposed.

We introduce \method, a training architecture that exploits this asymmetry without leaking privileged information into execution. The actor predicts a five-dimensional \emph{verifiable feasibility representation} (VFR): contact margin, energy difference, reachability, target utilization, and deadline buffer. Once the actor proposes an action, an environment verifier computes its exact action-conditioned target. VFR supervision makes the policy's feasibility estimate inspectable; an action-consistency objective connects that estimate to control. A \emph{counterfactual group-relative advantage} (CGRA) then scores a small group of candidate representation--action pairs against the same state and peer requests. Verification and candidate scoring are used only during training. At execution, every agent uses its local observation in a single forward pass.

The central scientific question is not whether additional shaping improves return, but which component creates the gain and when that gain transfers. We address this with reward-matched baselines, a seven-step mechanism ladder, a $4\times2$ representation--CGRA factorial, coordinate-removal and permutation controls, ten paired seeds, seven topology families, cross-topology calibration, scale sweeps, and three non-SAGIN stress tests. This design separates verifier grounding from reward choice, architectural capacity, and the relative-credit update.

The resulting evidence is both strong and mechanistically specific. Under identical execution rewards, \method improves success over reward-matched, attention-augmented Beta-distribution multi-agent proximal policy optimization (AB-MAPPO-R) by $8.74$ points and reduces coverage violation from $52.70\%$ to $0.51\%$ over ten paired seeds. Across four representation types, CGRA contributes $14.16$--$16.48$ success points, whereas the interaction with representation type is not detected. The supervised VFR is therefore an auditable feasibility interface; CGRA is the dominant optimization mechanism in SAGIN. Seven-topology evaluation, signed-margin calibration, and user/satellite scale sweeps show that this combination remains effective beyond the training layout. Resource-allocation and force/contact tests then reveal a complementary result: when the local candidate score conflicts with a shared constraint, the same credit mechanism exposes that mismatch through reproducible failure.

\paragraph{Contributions.}
\begin{itemize}[leftmargin=1.25em,itemsep=1.5pt,topsep=2pt]
    \item \textbf{Action-conditioned feasibility as a learnable interface.} We formulate VFRs, compact policy variables with exact post-action training targets but no privileged execution inputs, and connect their prediction error to feasibility-sign errors through a margin bound.
    \item \textbf{Counterfactual credit with explicit mechanism attribution.} We pair VFR with CGRA and isolate its effect using reward-matched controls, a seven-arm ladder, and a ten-seed factorial across handcrafted, learned, random, and latent representations.
    \item \textbf{Scale, transfer, and execution evidence.} \method reaches $55.33\pm3.60\%$ success with $0.45\pm0.81\%$ coverage violation, remains $24.33$--$30.02$ points above MAPPO-R on seven topologies, sustains $50.1$--$53.8\%$ success from 10 to 40 users, and adds $0.026$ ms to a central processing unit (CPU) actor step.
    \item \textbf{A measured applicability criterion.} Cross-domain factorials, collapse attribution, and pre-registered stabilizer tests show that CGRA is effective when its candidate score respects the environment's constraint geometry and predictably unsafe when local score improvement rewards shared-resource over-demand.
\end{itemize}

%% file: sections/related.tex
\section{Related Work}
\label{sec:related}
\paragraph{Cooperative MARL and counterfactual credit.}
Centralized training with decentralized execution (CTDE) addresses non-stationarity by allowing critics to use global information while actors retain local inputs. Representative approaches include multi-agent deep deterministic policy gradient (MADDPG), value factorization through QMIX, and multi-agent proximal policy optimization (MAPPO)~\cite{08,20,07,36,mat}. Sequence-modeling formulations cast offline cooperative control as return-conditioned next-action prediction~\cite{decision_transformer,madt}. Counterfactual multi-agent (COMA) policy gradients marginalize an agent's action to construct an agent-specific baseline, while factored multi-agent centralized policy gradients (FACMAC) optimize a factored centralized critic in continuous action spaces~\cite{47,49}. Heterogeneous-agent policy optimization provides monotonic-improvement analysis for sequential multi-agent updates~\cite{48,hasac,harl}. CGRA addresses a different credit object: it compares sampled representation--action candidates for one agent under fixed same-slot peer requests, then mixes their normalized relative score with generalized advantage estimation (GAE)~\cite{38}. It requires neither a privileged critic at execution nor enumeration of the joint action space.

\paragraph{Safe and constrained control.}
Constrained policy optimization and Lagrangian methods optimize expected return subject to expected costs~\cite{14,15,52,safe_rlhf,cup,saute_rl,safety_gymnasium,omnisafe,cal}. In networked systems, constrained MARL has been used for dynamic routing, while graph-based policies capture interference and resource-coupling structure~\cite{12,13,29,macpo}. These methods regulate long-run costs but do not necessarily expose why a particular action is feasible. Hard action masking provides per-decision enforcement when exact constraints are known, but that information is privileged in our decentralized benchmark. VFR instead predicts the missing action-conditioned quantities and makes prediction error, consistency, and signed margins directly measurable. It complements, rather than replaces, hard safety layers when such layers are available.

\paragraph{Decision-relevant and interpretable representations.}
State abstraction seeks compact representations that preserve optimal decisions~\cite{41}; causal, invariant, and bisimulation objectives aim to discard nuisance variation while retaining behaviorally relevant structure~\cite{42,43,44,45}. Concept bottleneck models use annotated intermediate variables to make predictions inspectable and editable~\cite{50,51}; recent work extends this idea to reinforcement-learning policies~\cite{46}, and feasibility-consistent objectives learn safety-aware embeddings for constrained control~\cite{fcsrl}. VFR shares their explicit semantics but differs in two respects. Its targets are computed from the realized state--action pair rather than attached to the observation alone, and the action head retains a backbone bypass. The representation is therefore a supervised, auditable pathway rather than a strict information bottleneck. Our strict-bottleneck and pathway interventions measure the consequence of that choice.

\paragraph{Learning for SAGIN resource management.}
Space--air--ground integrated networking couples orbital contact, wireless access, edge computation, and shared capacity~\cite{03,05,18}. Deep reinforcement learning has been applied to mobile-edge offloading, satellite edge computing, hybrid cloud--edge assignment, SAGIN traffic control, UAV-enabled computation, graph-based allocation, and satellite routing~\cite{09,21,22,23,24,25,27,28,29}. Most of this literature evaluates task reward under a fixed simulator and reports aggregate constraints. Our focus is complementary: learning an explicit action-conditioned feasibility interface, identifying its relationship to credit assignment, and testing the resulting mechanism across topology, scale, and reward geometry.

\paragraph{Verifier supervision and group-relative optimization.}
Outcome and process verifiers provide dense supervision when an environment can check intermediate outputs~\cite{30}. Group-relative policy optimization normalizes scores within sampled groups to avoid a separate value model~\cite{31}, while preference-based objectives replace absolute rewards with relative comparisons~\cite{dpo}. \method adapts these ideas to continuous physical control: the verifier returns typed feasibility margins rather than textual correctness, and the group varies a single agent's representation--action candidate while holding peer requests fixed. The reward-matched factorial is essential here because it separates the benefit of relative credit from the semantic content of the representation.

%% file: sections/method.tex
\section{Verifiable Feasibility and Counterfactual Credit}
\label{sec:method}
\subsection{Constrained decentralized control}
We consider a cooperative partially observable Markov game with agents $i\in\{1,\ldots,K\}$. The global state $s_t$ contains task descriptors, agent positions, remote-node loads, link states, and moving-node visibility. Agent $i$ receives a local observation $o_{i,t}\in\R^{d_o}$ and chooses $a_{i,t}=(\alpha_{i,t},\rho_{i,t},B_{i,t})$: an execution target $\alpha_i$, offloading ratio $\rho_i\in[0,1]$, and bandwidth request $B_i\in[0,1]$. A task succeeds only if end-to-end latency is below deadline $D_{\max}$ and the selected remote target remains reachable for the transmission interval. Shared bandwidth and compute introduce coupling among otherwise decentralized actions. Appendix~\ref{app:env} gives the physical model.

The training simulator observes $s_t$ and can evaluate the realized action, but execution follows CTDE: $\pi_i$ receives only $o_{i,t}$. This creates an information asymmetry that \method uses as supervision. The training-time verifier may reveal whether a proposed action is feasible; the execution policy must predict that answer before acting.

\subsection{Verifiable feasibility representation}
A shared encoder computes $h_i=f_\theta(o_i)$. A representation head predicts
\begin{equation}
\chat_i=\phi_\theta(h_i)=
\left[\hat c_i^{cm},\hat c_i^{ed},\hat c_i^{cv},\hat c_i^{cu},\hat c_i^{db}\right]\in\R^5,
\label{eq:vfr}
\end{equation}
and the actor samples
\begin{equation}
a_i\sim\pi_\theta(\cdot\mid h_i,\chat_i).
\label{eq:actor}
\end{equation}
Table~\ref{tab:vfr_coordinates} summarizes the five coordinates. Each coordinate is normalized to a fixed numerical range before supervision. The direct $h_i$ pathway is retained deliberately: VFR is an explicit, grounded pathway that can be inspected and intervened on, not a claim that five variables are a sufficient statistic for optimal control.

\begin{table}[t]
\centering
\caption{Coordinates of the verifiable feasibility representation. Every target is computed for the action already proposed by the actor.}
\label{tab:vfr_coordinates}
\small
\setlength{\tabcolsep}{3.5pt}
\begin{tabular}{lll}
\toprule
Coordinate & Physical quantity & Feasibility information\\
\midrule
$c^{cm}$ & contact margin & remaining contact versus transmission time\\
$c^{ed}$ & energy difference & selected offload versus local execution\\
$c^{cv}$ & coverage validity & binary reachability of the selected target\\
$c^{cu}$ & compute utilization & post-action load at the selected target\\
$c^{db}$ & deadline buffer & deadline minus predicted completion time\\
\bottomrule
\end{tabular}
\end{table}

\begin{figure}[t]
\centering
\includegraphics[width=0.8\linewidth]{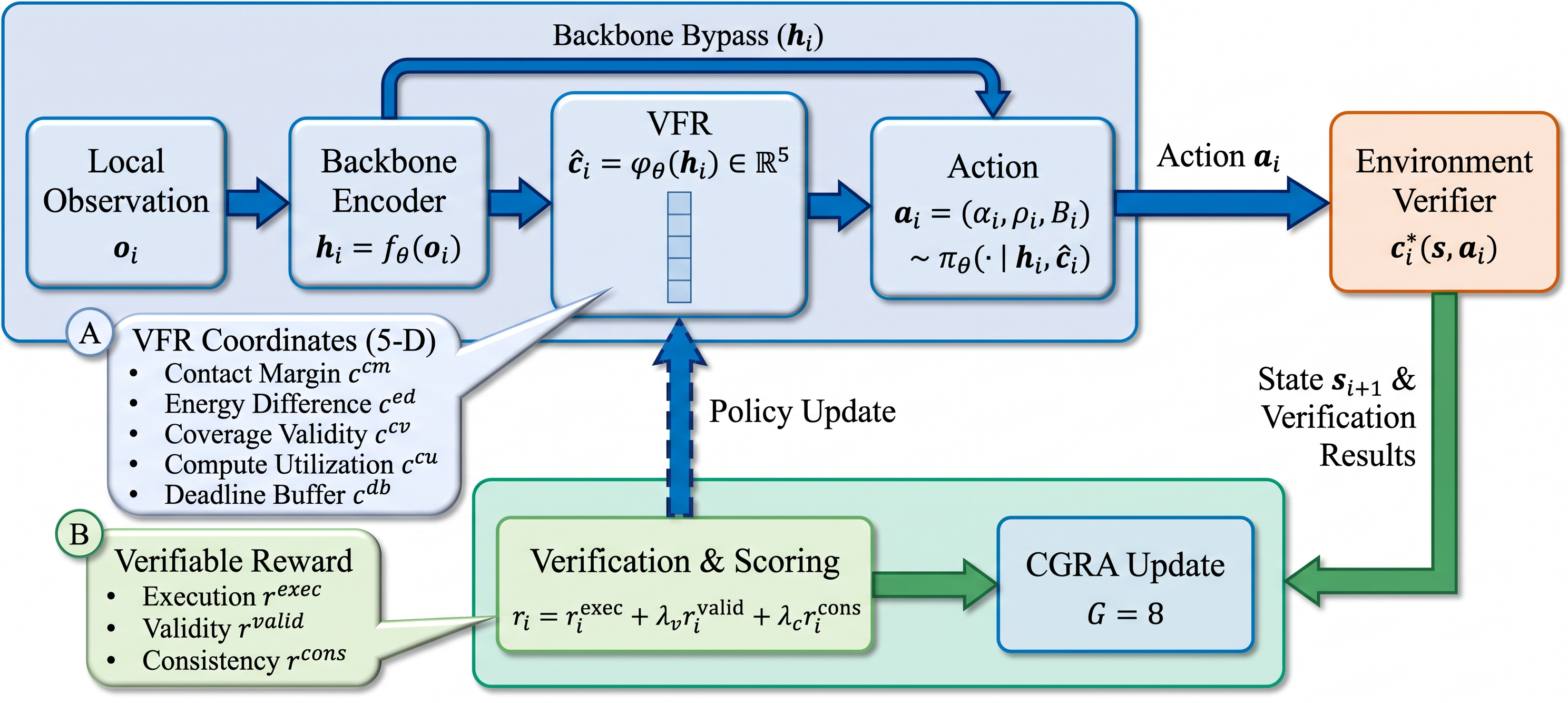}
\caption{\textbf{\method training and execution.} The actor predicts VFR and conditions its action on both VFR and backbone features. The environment computes the exact target only after an action is proposed. The green verification and candidate-scoring path is removed at execution.}
\label{fig:vera}
\end{figure}
Figure~\ref{fig:vera} resolves a potential ordering ambiguity. The actor first predicts $\chat_i$ and samples $a_i$ through Eq.~\eqref{eq:actor}; only then does the environment compute $\cgt_i(s,a_i)$. No action-conditioned target is available to the actor before its decision.

\paragraph{Feasibility-margin interpretation.}
Let $\bm m(s,a)\in\R^q$ collect signed feasibility margins, where $m_j(s,a)\ge0$ denotes satisfaction of constraint $j$, and let $\hat{\bm m}$ denote their prediction before thresholding. For any $\epsilon>0$,
\begin{equation}
\Pr[\exists j:\operatorname{sign}(\hat m_j)\ne \operatorname{sign}(m_j)]
\le
\Pr[\|\hat{\bm m}-\bm m\|_\infty>\epsilon]
+
\Pr[\min_j |m_j|\le\epsilon].
\label{eq:margin_bound}
\end{equation}
When prediction error is at most $\epsilon$ and every true margin lies more than $\epsilon$ from its boundary, no sign can flip. The bound does not imply optimal control; it identifies the two measurable sources of a wrong feasibility verdict: representation error and operation near a physical boundary. Appendix~\ref{app:margin} gives the proof and the binary-coverage construction.

\subsection{Action-conditioned verification}
VFR is trained with a coordinate-wise smooth-$\ell_1$ loss
\begin{equation}
\mathcal L_{\mathrm{VFR}}(\theta)=
\frac{1}{5}\sum_{j=1}^{5}\operatorname{smooth}\ell_1(\hat c_{i,j},c^\star_{i,j}).
\end{equation}
We also expose verification to the policy objective through validity and consistency terms,
\begin{equation}
r_i^{\mathrm{valid}}=-\|\chat_i-\cgt_i\|_1,\qquad
r_i^{\mathrm{cons}}=-\operatorname{viol}(a_i,\chat_i),
\end{equation}
where $\operatorname{viol}$ measures whether the selected target, offloading ratio, and request contradict the predicted margins. The complete reward is
\begin{equation}
r_i=r_i^{\mathrm{exec}}+\lambda_v r_i^{\mathrm{valid}}+\lambda_c r_i^{\mathrm{cons}}.
\label{eq:vera_reward}
\end{equation}
The complete configuration uses stronger success shaping and an explicit coverage penalty in $r_i^{\mathrm{exec}}$. We therefore do not attribute Table~\ref{tab:primary} to architecture alone. Reward-matched baselines receive the identical execution reward, and the mechanism ladder fixes that reward while varying supervision, conditioning, consistency, bottlenecking, and CGRA.

\subsection{Counterfactual group-relative advantage}
For each state--agent pair, the actor samples $G=8$ representation--action candidates. A one-sided what-if evaluator scores candidate $g$ while keeping peers' realized same-slot requests fixed, producing $r_i^{(g)}$. With group mean $\mu_i$ and standard deviation $\sigma_i$, the executed candidate receives
\begin{equation}
A_i^{\mathrm{grp}}=
\operatorname{clip}\!\left(
\frac{r_i^{(0)}-\mu_i}{\sigma_i+\epsilon},-3,3
\right).
\label{eq:cgra}
\end{equation}
The policy update mixes this term with the ordinary GAE advantage,
\begin{equation}
A_i=A_i^{\mathrm{GAE}}+\lambda_g A_i^{\mathrm{grp}},
\end{equation}
inside the clipped proximal policy optimization objective~\cite{38,39}. We call Eq.~\eqref{eq:cgra} the \emph{counterfactual group-relative advantage}. It is inspired by group-relative normalization~\cite{31,dr_grpo}, but it is neither a joint counterfactual oracle nor COMA's marginalized baseline: peer reactions and future dynamics are not re-simulated. This distinction explains both its efficiency and its applicability condition. CGRA is informative when fixed-peer candidate scores preserve the relevant constraint ordering; it can be biased when an agent's locally favorable request creates an unpriced externality.

\paragraph{Training and execution complexity.}
Training evaluates $G$ candidates per selected state--agent pair, whereas execution computes only $f_\theta$, $\phi_\theta$, and $\pi_\theta$ once. The verifier, counterfactual evaluator, global state, and peer requests are absent from the deployed path. Measured latency in Sec.~\ref{sec:scale} confirms that this separation leaves the decentralized actor within $0.055$ ms of MAPPO on both tested devices.

%% file: sections/experiments.tex
\section{Experiments}
\label{sec:exp}
\paragraph{Environment, baselines, and reporting.}
The standard SAGIN has 20 mobile agents, six LEO satellites at 550 km, and four UAV nodes; local observation and action dimensions are 66 and 13. Unless stated otherwise, training lasts 3,000 episodes and each seed is summarized over its final 500 episodes (the scale sweeps use a 500-episode budget and report the converged window, episodes $>$ 250). The primary benchmark uses five seeds $S5=\{42,123,2025,7,2024\}$; confirmatory studies add $\{11,19,73,314,2718\}$ to form ten paired seeds (S10). We report mean$\pm$sample standard deviation, two-sided paired tests for matched designs, paired standardized effect $d_z$, bootstrap confidence intervals, and Holm correction for topology families. Baselines include MAPPO~\cite{07}, Beta-distribution MAPPO (B-MAPPO), AB-MAPPO, MADDPG~\cite{08}, a Lagrangian constrained policy~\cite{14}, and a privileged hard-mask reference. Vectorized Multi-Agent Simulator (VMAS) navigation~\cite{vmas} supplies an independent control domain. Appendix~\ref{app:protocol} gives all metrics, hyperparameters, and test definitions.

\subsection{Primary performance and constraint compliance}
Table~\ref{tab:primary} separates the standard complete-method benchmark from the stronger reward-matched comparison. In the standard S5 protocol, \method is the best non-privileged method in success rate (SR), energy, Jain's fairness index, and coverage violation (COV), reaching $55.33\pm3.60\%$ SR with $0.45\pm0.81\%$ COV. Its success lies $1.33$ points below MaskMAPPO, which receives an exact feasibility mask at decision time. The complete standard metrics and the eight-method comparison appear in Appendix Table~\ref{tab:app_main} and Fig.~\ref{fig:app_main_comparison}.

\begin{table}[t]
\centering
\caption{Primary SAGIN results. Values are mean$\pm$sample standard deviation across seeds. Standard S5 and reward-matched S10 are separate protocols; MaskMAPPO$^\dagger$ observes exact decision-time feasibility.}
\label{tab:primary}
\small
\setlength{\tabcolsep}{4.0pt}
\begin{tabular}{llrr}
\toprule
Protocol & Method & SR (\%) $\uparrow$ & COV (\%) $\downarrow$\\
\midrule
\multirow{5}{*}{Standard S5}
 & \textbf{\method} & \textbf{55.33$\pm$3.60} & \textbf{0.45$\pm$0.81}\\
 & AB-MAPPO & 50.68$\pm$4.65 & 45.32$\pm$9.42\\
 & MAPPO & 26.35$\pm$6.15 & 69.54$\pm$11.07\\
 & Tuned constrained policy & 40.17$\pm$3.46 & 33.65$\pm$8.67\\
 & MaskMAPPO$^\dagger$ & 56.66$\pm$0.70 & 0.05\\
\midrule
\multirow{2}{*}{Reward-matched S10}
 & \textbf{\method} & \textbf{56.00$\pm$5.19} & \textbf{0.51$\pm$0.71}\\
 & AB-MAPPO-R & 47.26$\pm$6.34 & 52.70$\pm$10.55\\
\bottomrule
\end{tabular}
\end{table}

Because the complete \method configuration also changes the execution reward, the standard block alone cannot identify an architectural effect. In the reward-matched S10 protocol, both methods receive the same execution reward and budget. \method improves SR by $8.74$ points ($d_z=0.864$, $p=0.0231$, bootstrap 95\% confidence interval $[3.20,14.97]$), with a positive difference on 8/10 seeds. It reduces COV by $52.19$ points ($d_z=-5.13$, $p=5.7\times10^{-8}$, interval $[-58.13,-46.24]$), improving compliance on every seed. Exact sign-flip tests give $p=0.0117$ and $p=0.0020$. A nine-setting constrained-policy sweep reaches $40.17\pm3.46\%$ SR and $33.65\pm8.67\%$ COV, showing that the gap is not explained by an untuned penalty baseline.

\subsection{Mechanism attribution}
\label{sec:mechanism}
We first fix the execution reward and retrain seven variants (Appendix Table~\ref{tab:ladder} and Fig.~\ref{fig:app_ladder}). Supervised auxiliary VFR reduces COV relative to an ungrounded five-dimensional latent head from $25.56\%$ to $2.40\%$ ($p=0.042$). Feeding VFR directly to the actor adds no detected gain over auxiliary supervision. Removing CGRA, however, reduces SR from $54.57\%$ to $40.12\%$ ($p=0.0023$) and raises COV from $2.33\%$ to $62.23\%$ ($p<10^{-4}$). A retrained strict bottleneck also underperforms the complete actor, confirming that backbone access is a useful design choice rather than an accidental shortcut.

The S10 $4\times2$ factorial makes this attribution independent of a single representation design (Fig.~\ref{fig:mechanism}a; Appendix Table~\ref{tab:u4}). Across handcrafted, learned, random-supervised, and latent heads, CGRA adds $14.91$, $16.06$, $16.48$, and $14.16$ SR points; every paired test has $p\le6.7\times10^{-6}$. COV changes range from $-42.08$ to $-62.13$ points. A repeated-measures Friedman test detects no difference among the four SR effects ($p=0.323$). Coordinate removal, random-target, and permutation controls change SR by at most $1.74$ points (Appendix Table~\ref{tab:semantic_controls} and Fig.~\ref{fig:semantic_controls}). Collectively, these experiments identify CGRA as the main performance driver and VFR as the grounded interface that makes feasibility error and consistency auditable.

\subsection{Topology transfer and feasibility calibration}
\begin{figure}[t]
\centering
\includegraphics[width=0.80\linewidth]{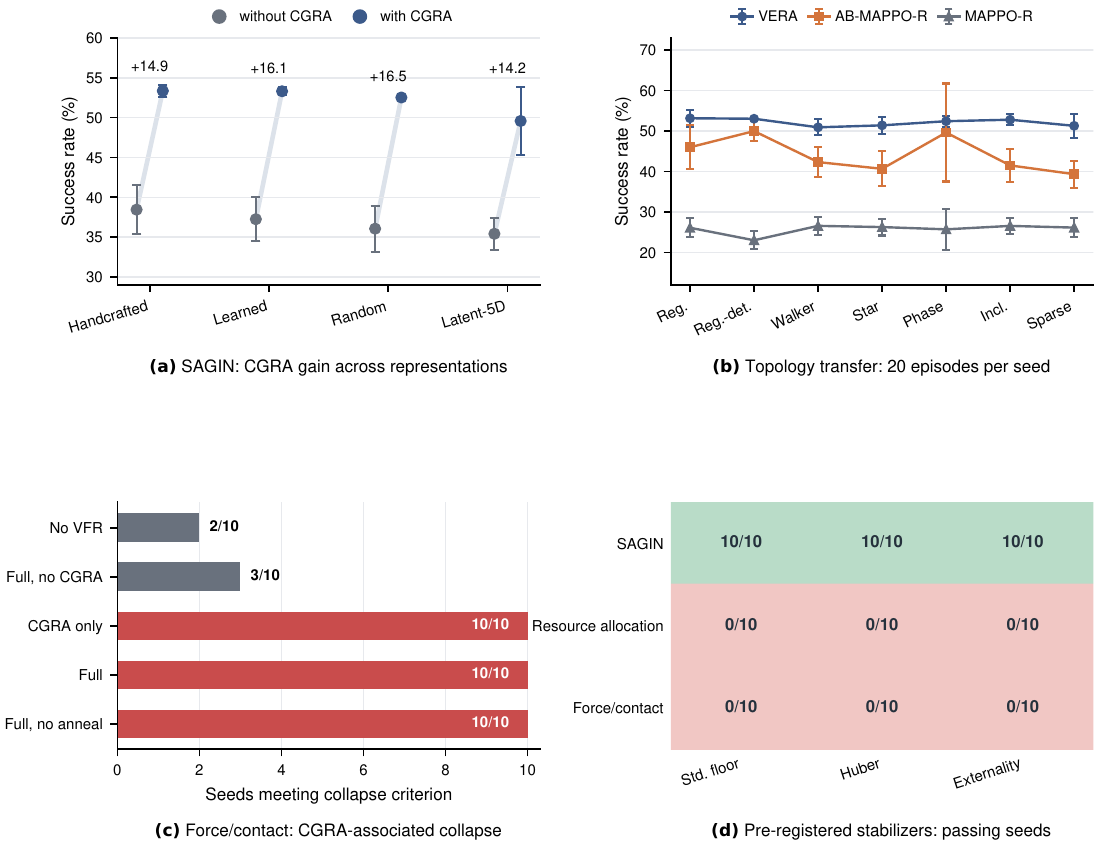}
\caption{\textbf{Mechanism and scope.} (a) CGRA improves S10 SAGIN success across four representations. (b) \method transfers across seven topologies; intervals use seed-level 95\% confidence intervals from 20 episodes per seed. (c) All CGRA-bearing force/contact arms meet the pre-registered collapse rule. (d) Three stabilizers preserve SAGIN but pass neither cross-domain rescue gate. Standard-deviation bars are used in (a).}
\label{fig:mechanism}
\end{figure}

The topology study evaluates 20 episodes for every seed--method--topology cell across seven unseen layouts, mirroring the procedural-generalization principle behind SMACv2~\cite{smacv2}. \method remains at $50.93$--$53.17\%$ SR, compared with $23.03$--$26.59\%$ for reward-matched MAPPO (MAPPO-R) and $39.36$--$50.00\%$ for AB-MAPPO-R (Fig.~\ref{fig:mechanism}b). The paired \method--MAPPO-R gap is $24.33$--$30.02$ points and remains significant after Holm correction on all seven topologies. The \method--AB-MAPPO-R difference is positive on all seven ($2.80$--$11.95$ points), although those individual comparisons do not survive family-wise correction.

Transfer follows the physical margin rather than a topology label. The calibration study evaluates 80,118 decisions on each of four shifted topologies and bins the minimum coverage/deadline margin. All three clearly negative bins have $100\%$ violation, the boundary bin has $52.4$--$86.7\%$, and all four positive bins have $0\%$. Figure~\ref{fig:app_margin_calibration} and Appendix~\ref{app:calibration} show the simultaneous decline in VFR error. This ordered transition is the empirical counterpart of Eq.~\eqref{eq:margin_bound}.

\subsection{Scaling and execution profile}
\label{sec:scale}
\begin{figure}[t]
\centering
\includegraphics[width=0.85\linewidth]{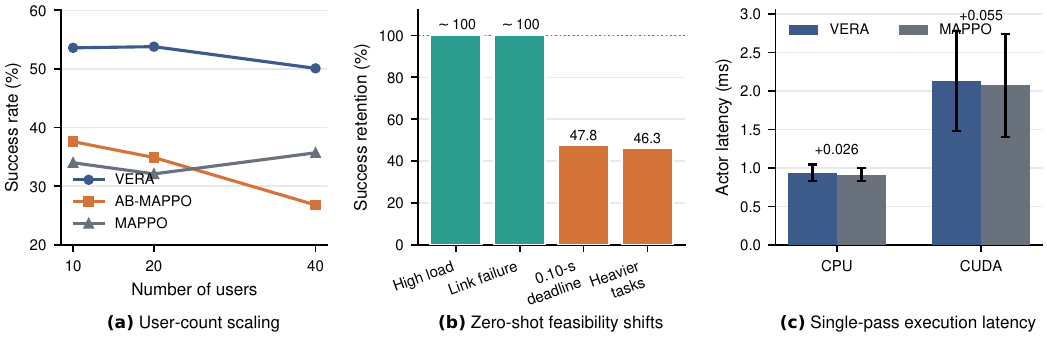}
\caption{\textbf{Scale, zero-shot shifts, and execution cost.} (a) Mean SR as the number of users grows. (b) \method SR retention under dimension-preserving shifts; the tighter deadline and heavier tasks also reduce intrinsic feasibility. (c) Per-step actor latency, reported as mean$\pm$sample standard deviation. Panel titles are placed below each subplot.}
\label{fig:scale_deployment}
\end{figure}

The user-count sweep increases $K$ from 10 to 40 while keeping six satellites. \method sustains $53.6/53.8/50.1\%$ SR, whereas AB-MAPPO falls from $37.6\%$ to $26.8\%$ and MAPPO remains at $32.1$--$35.7\%$ (Fig.~\ref{fig:scale_deployment}a). With $K=20$ and 3/6/12 satellites, \method obtains $55.1/53.8/53.0\%$ SR; COV remains below $11\%$ over all six scale settings. Without retraining, high load and link failure retain $100\%$ of source SR, while a $0.10$-s deadline and heavier tasks retain $47.8\%$ and $46.3\%$ as the feasible task set contracts (Fig.~\ref{fig:scale_deployment}b).

Execution uses one actor pass. Central processing unit (CPU) latency is $0.939\pm0.106$ ms for \method and $0.913\pm0.086$ ms for MAPPO; Compute Unified Device Architecture (CUDA) latency is $2.128\pm0.649$ and $2.073\pm0.668$ ms, respectively (Fig.~\ref{fig:scale_deployment}c). Thus the VFR pathway adds $0.026$ ms on CPU and $0.055$ ms on CUDA. The $G=8$ candidate scorer costs 6.94 ms on CPU and 8.81 ms on CUDA per update, but is absent from execution.

\subsection{Reward geometry determines cross-domain behavior}
VMAS cooperative navigation transfers the training pattern but not the superiority of fixed semantics. Full \method reaches $24.75\pm4.09\%$ goal success versus $4.86\pm4.20\%$ without VFR/CGRA ($p=0.0020$); a Latent-5D head reaches $28.41\pm3.69\%$ and exceeds Full by $3.66$ points ($p=0.038$). The 4,000-episode out-of-distribution (OOD) audit similarly shows a goal--collision trade-off rather than universal dominance (Appendix~\ref{app:vmas} and Fig.~\ref{fig:boundary}c--d).

Two stress domains isolate the failure mechanism. In resource allocation, every CGRA arm converges near $63.47\%$ SR but violates shared capacity on $100\%$ of decisions; without CGRA, SR is $49.27$--$49.73\%$ and COV is $1.43$--$5.70\%$ across all four representations (Fig.~\ref{fig:boundary}a--b). The fixed-peer evaluator rewards over-demand because the local score omits the shared externality. In force/contact control, CGRA-only, Full, and Full without annealing collapse on 10/10 seeds, compared with 2/10 and 3/10 for the two CGRA-absent arms (Fig.~\ref{fig:mechanism}c). Standard-deviation flooring, Huberized relative advantage, and an externality correction each preserve SAGIN on 10/10 seeds, yet pass neither rescue gate (Fig.~\ref{fig:mechanism}d). These factorial and pre-registered tests turn transfer failure into a concrete design criterion: candidate evaluation must price the constraints that an individual action externalizes.

%% file: sections/conclusion.tex
\section{Conclusion}
\label{sec:conclusion}
\method separates two signals that end-to-end constrained policies commonly conflate: the return associated with an action and the physical reason that action can execute. Its VFR converts exact post-action feasibility calculations into a compact supervised interface, while CGRA converts within-state candidate comparisons into agent-specific credit. Neither training mechanism appears in the decentralized execution path.

The evidence supports three conclusions. First, the combined method is effective in dynamic SAGIN control: it reaches $55.33\pm3.60\%$ SR with $0.45\pm0.81\%$ COV, gains $8.74$ SR points over the strongest reward-matched baseline on ten paired seeds, and reduces that baseline's violation by $52.19$ points. Second, the factorial identifies where the gain enters. CGRA contributes $14.16$--$16.48$ points across four representations, while VFR supplies the verifier-grounded interface through which margin error, action consistency, and topology transfer can be audited. Third, the method retains a $24.33$--$30.02$ point gap over MAPPO-R on seven layouts, sustains $50.1$--$53.8\%$ SR from 10 to 40 users with coverage violation below $11\%$ over all six scale settings, and adds only $0.026$ ms to a CPU actor step.
The cross-domain results sharpen rather than dilute the contribution. CGRA succeeds when fixed-peer candidate scores preserve the ordering induced by the relevant physical constraints; it fails when locally favorable actions impose an unpriced shared externality. This criterion explains both the compliant SAGIN result and the resource/force failures, and it is testable before deployment through the same factorial and stress-test protocol. More broadly, \method shows how simulator-accessible physics can supervise an auditable policy interface without becoming a privileged execution input. Extending the evaluator to learn or price constraint geometry is the natural next step toward general-purpose verifiable multi-agent control.
\FloatBarrier

%% file: sections/appendix.tex
\section{Extended Experimental and Implementation Details}
\label{app:details}

This appendix is part of the preprint portable document format (PDF) file and provides the physical model, objectives, complete protocols, secondary analyses, and experiment-to-artifact map needed to audit the main claims. All plots use the final ten-seed, 20-episode, or 4,000-record analyses where those higher-precision evaluations are available.

\subsection{Environment and physical model}
\label{app:env}
The benchmark is a three-tier space--air--ground integrated network with $K$ ground users, $M_{sat}$ LEO satellites, and $M_{uav}$ UAV nodes. The standard configuration is $K=20$, $M_{sat}=6$, and $M_{uav}=4$. LEO altitude is 550 km, UAV altitude is 20 km, the episode horizon is 200 slots, and the minimum satellite elevation is $10^\circ$. Satellite angular velocity is $\omega=\sqrt{\mu/r^3}$ for $r=R_E+h_{sat}$, corresponding to an approximately 95-minute orbit. Visibility ends when ground-projected distance exceeds the elevation-implied threshold.

At each slot, agent $k$ receives a task with probability 0.8. Task size is sampled from $[5\times10^4,2\times10^5]$ bits and compute intensity from $[500,1500]$ cycles/bit. The standard deadline is $D_{\max}=0.15$ s. For target $n$, the transmission rate is
\begin{equation}
R_{k,n}(t)=B_{k,n}(t)\log_2\!\left(1+\frac{P_k g_{k,n}(t)}{N_0B_{k,n}(t)}\right).
\end{equation}
For offloading ratio $\rho_k$ and target $\alpha_k$, remote time is
\begin{equation}
T_k^{off}=\frac{\rho_k L_k}{R_{k,\alpha_k}}+\frac{\rho_k L_k C_k}{f_{\alpha_k}},
\end{equation}
local time is $T_k^{loc}=(1-\rho_k)L_kC_k/f_k^{loc}$, and task latency is $T_k=\max(T_k^{loc},T_k^{off})$. A remote task succeeds only if its link remains visible for the required transmission interval. Concurrent bandwidth requests are allocated proportionally, so shared-resource contention is explicit. The simulator uses analytic orbital motion and a regional ground-track abstraction; it omits explicit Doppler re-synchronization and inter-cell interference.

\subsection{Feasibility-margin bound}
\label{app:margin}
For Eq.~\eqref{eq:margin_bound}, let $E=\{\|\hat{\bm m}-\bm m\|_\infty\le\epsilon\}$ and $B=\{\min_j|m_j|>\epsilon\}$. On $E\cap B$, $|\hat m_j-m_j|\le\epsilon<|m_j|$ for every coordinate; hence no prediction crosses zero and every feasibility sign is preserved. Any sign mismatch is therefore contained in $E^c\cup B^c$, and the union bound gives Eq.~\eqref{eq:margin_bound}. The binary coverage-validity entry in the implementation is obtained by thresholding a signed contact-time margin, so the statement applies to its pre-threshold value.

\subsection{Rewards, protocol, and statistics}
\label{app:protocol}
\label{app:reward}
Ordinary baselines use
\begin{equation}
r^{base}=0.5(10SR)+0.25\frac{5}{1+T}+0.25\,5\max(0,1-E/E_{th}).
\end{equation}
VERA's complete execution layer uses nonlinear success shaping, threshold bonuses, latency and energy terms, and a squared coverage penalty,
\begin{align}
r^{exec}_{VERA}={}&\lambda_{sr}\big(20SR^{1.2}+b(SR)\big)+\lambda_T r_T+\lambda_E r_E\\
&-10\lambda_{cov}v_{cov}^2,
\end{align}
followed by validity and consistency rewards. The complete-method table is therefore not an architecture-only comparison. Reward-matched controls receive the same execution reward; the mechanism ladder further holds that reward fixed while varying VFR supervision, conditioning, consistency, bottlenecking, and CGRA.

All proximal-policy-optimization-family methods use centralized training and decentralized execution, Kullback--Leibler (KL) early stopping, learning-rate and entropy annealing, and value clipping. The primary five-seed set is $S5=\{42,123,2025,7,2024\}$. Confirmatory S10 results add $\{11,19,73,314,2718\}$. Unless a subsection states otherwise, training runs for 3,000 episodes and the final 500 episodes form each seed-level estimate (the scale sweeps use a 500-episode budget and report the converged window, episodes $>$ 250). We report mean$\pm$sample standard deviation across seeds. Matched designs use two-sided paired $t$ tests, paired standardized effect $d_z$, paired bootstrap intervals, and exact sign-flip tests where reported. We do not interpret $p>0.05$ as equivalence. The topology study uses 20 evaluation episodes per seed; the VMAS OOD study is an inference-only audit on the five new seeds with 100 episodes for every variant--shift cell, not a ten-seed training replication.

\subsection{Complete standard-configuration results}
\label{app:main}
\begin{table}[!htbp]
\centering
\caption{Standard configuration, final-500-episode window, five seeds. Latency is averaged over successful tasks and must be read with success rate.}
\label{tab:app_main}
\small
\resizebox{\linewidth}{!}{%
\begin{tabular}{lrrrrr}
\toprule
Method & SR (\%) & Latency (s) & Energy & Cov. viol. (\%) & Jain\\
\midrule
VERA & 55.33$\pm$3.60 & 0.1010$\pm$0.0003 & 0.018$\pm$0.015 & 0.45$\pm$0.81 & 0.9936$\pm$0.0007\\
AB-MAPPO & 50.68$\pm$4.65 & 0.0956$\pm$0.0028 & 8.482$\pm$2.689 & 45.32$\pm$9.42 & 0.9825$\pm$0.0092\\
B-MAPPO & 49.14$\pm$5.39 & 0.0950$\pm$0.0040 & 11.849$\pm$10.424 & 46.65$\pm$10.16 & 0.9841$\pm$0.0065\\
MAPPO & 26.35$\pm$6.15 & 0.0874$\pm$0.0027 & 55.851$\pm$40.247 & 69.54$\pm$11.07 & 0.9779$\pm$0.0057\\
MADDPG & 21.10$\pm$6.63 & 0.0722$\pm$0.0076 & 6.219$\pm$1.951 & 91.74$\pm$10.19 & 0.8893$\pm$0.0335\\
Random & 14.86$\pm$1.51 & 0.0829$\pm$0.0025 & 179.987$\pm$13.650 & 97.17$\pm$1.95 & 0.9447$\pm$0.0154\\
MaskMAPPO & 56.66$\pm$0.70 & -- & -- & 0.05 & --\\
Tuned ConstrainedMAPPO & 40.17$\pm$3.46 & -- & -- & 33.65$\pm$8.67 & --\\
\bottomrule
\end{tabular}%
}
\end{table}
\FloatBarrier

\begin{figure}[!htbp]
\centering
\includegraphics[width=0.91\linewidth]{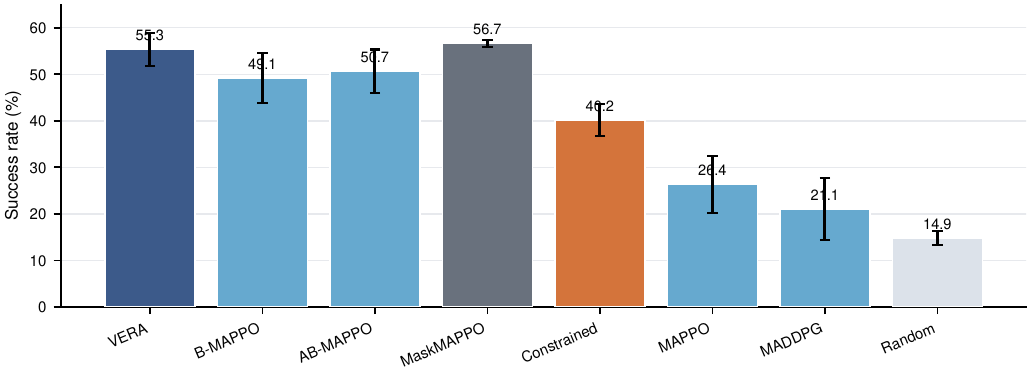}
\caption{Standard-configuration success across the complete method comparison. MaskMAPPO uses a privileged hard feasibility mask and is shown as a reference rather than a deployable competitor; the constrained baseline is the tuned configuration from Table~\ref{tab:primary}. Error bars are sample standard deviations across five seeds.}
\label{fig:app_main_comparison}
\end{figure}

Table~\ref{tab:app_main} reports every standard metric, and Fig.~\ref{fig:app_main_comparison} visualizes the success comparison.
\FloatBarrier

The paired Full--w/o-VFR same-objective comparison gives $+2.58$ success points ($p=0.218$, $d_z=0.65$) and $-10.31$ violation points ($p=0.060$, $d_z=-1.17$). A 200,000-replicate paired bootstrap gives intervals $[-1.0,5.3]$ and $[-17.1,-3.4]$ points; exact five-pair sign-flip $p$ values are 0.25 and 0.125. These coarse five-seed tests motivate the larger reward-matched and factorial studies rather than supporting a strong mean claim by themselves.

\subsection{Same-objective ablation and zero-shot shifts}
\label{app:ablation_shifts}
Table~\ref{tab:app_ablation} isolates VFR conditioning, the coverage coordinate, and the verification reward before the larger mechanism ladder and factorial tests:
\begin{table}[!htbp]
\centering
\caption{Five-seed ablation under the VERA execution objective.}
\label{tab:app_ablation}
\small
\begin{tabular}{lrrr}
\toprule
Variant & SR (\%) & Cov. viol. (\%) & VFR validity (\%)\\
\midrule
Full & 54.57$\pm$2.26 & 2.33$\pm$1.44 & 82.63$\pm$2.73\\
w/o VFR & 51.98$\pm$3.00 & 12.65$\pm$9.03 & --\\
w/o coverage coordinate & 53.45$\pm$0.51 & 0.78$\pm$0.65 & 87.39$\pm$0.83\\
w/o verification reward & 37.10$\pm$6.07 & 60.67$\pm$13.16 & 54.93$\pm$5.35\\
\bottomrule
\end{tabular}
\end{table}

\begin{figure}[!htbp]
\centering
\includegraphics[width=0.99\linewidth]{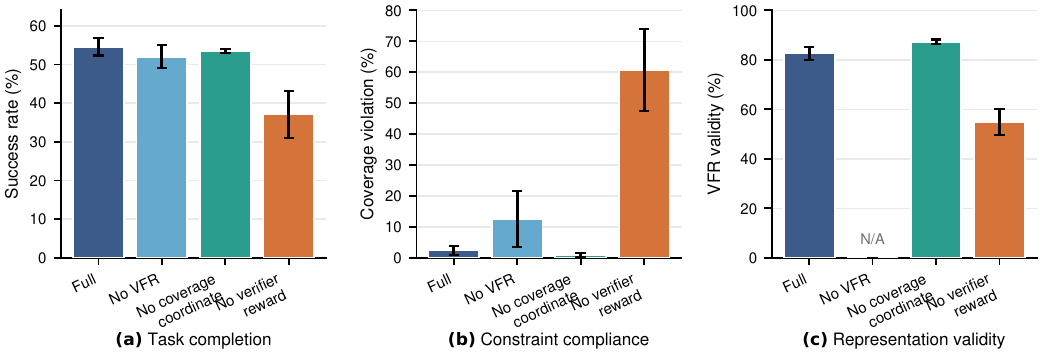}
\caption{Same-objective ablation of success, coverage violation, and representation validity. The larger reward-matched ladder and S10 factorial provide the stronger mechanism tests.}
\label{fig:app_ablation}
\end{figure}

Figure~\ref{fig:app_ablation} shows all three ablation metrics with a common visual style.
\FloatBarrier

With trained weights fixed, we also test high load, link failure, a tightened $0.10$-s deadline, and heavier tasks. VERA retains 100\% of source success under high load and link failure, but 47.8\% and 46.3\% under the tighter deadline and heavier tasks. The latter shifts also increase the fraction of structurally infeasible tasks. Figure~\ref{fig:scale_deployment}b summarizes these shifts.

\subsection{Ten-seed reward matching and constrained-baseline tuning}
\label{app:rewardmatched}
Table~\ref{tab:rewardmatched} repeats the reward-matched means alongside the paired analysis below.
\begin{table}[!htbp]
\centering
\caption{Reward-matched SAGIN comparison over ten paired seeds.}
\label{tab:rewardmatched}
\small
\begin{tabular}{lrr}
\toprule
Method & SR (\%) & Cov. viol. (\%)\\
\midrule
VERA & 56.00$\pm$5.19 & 0.51$\pm$0.71\\
AB-MAPPO-R & 47.26$\pm$6.34 & 52.70$\pm$10.55\\
\bottomrule
\end{tabular}
\end{table}
The paired differences are $+8.74$ success points ($d_z=0.864$, $p=0.0231$, bootstrap 95\% CI $[3.20,14.97]$) and $-52.19$ violation points ($d_z=-5.13$, $p=5.7\times10^{-8}$, CI $[-58.13,-46.24]$). Exact sign-flip $p$ values are 0.0117 and 0.0020. VERA improves success on 8/10 seeds and compliance on 10/10.

ConstrainedMAPPO is tuned over nine combinations of dual learning rate, constraint budget, and penalty initialization. The selected setting ($\mathrm{lr}_{\lambda}=0.3$, budget $0.06$, initialization $1.5$) is rerun on S5 and produces the value in Table~\ref{tab:primary}; selection and final evaluation are stored separately.

\subsection{Mechanism ladder, semantic controls, and S10 factorial}
\label{app:semantics}
\begin{figure}[!htbp]
\centering
\includegraphics[width=0.99\linewidth]{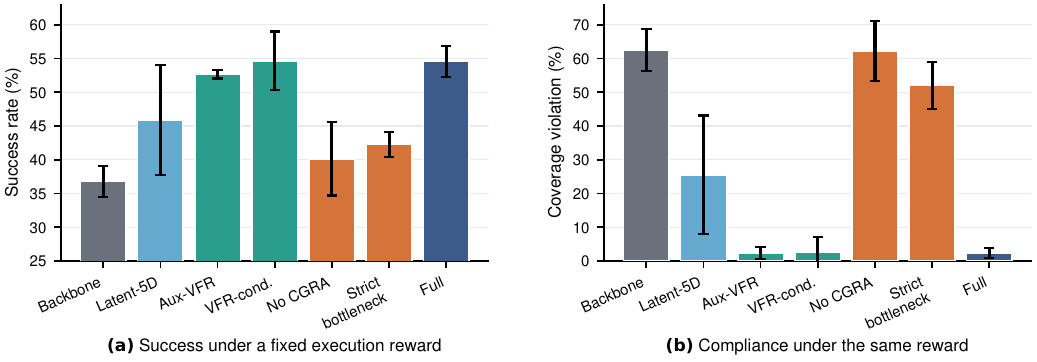}
\caption{Seven-way reward-matched mechanism ladder. (a) Converged success. (b) Coverage violation. Error bars are sample standard deviations across five seeds; panel titles appear below the subplots.}
\label{fig:app_ladder}
\end{figure}

The reward-matched mechanism ladder uses the switches in Table~\ref{tab:ladder_design}.
\begin{table}[!htbp]
\centering
\caption{Component switches in the reward-matched mechanism ladder.}
\label{tab:ladder_design}
\scriptsize
\begin{tabular}{lcccc}
\toprule
Variant & VFR supervision & VFR enters action & Consistency & CGRA\\
\midrule
Backbone & no & no & no & no\\
Latent-5D & no & yes & no & yes\\
Aux-VFR & yes & no & no & yes\\
VFR-cond. & yes & yes & no & yes\\
w/o CGRA & yes & yes & yes & no\\
Strict bottleneck & yes & yes, no backbone bypass & yes & no\\
Full & yes & yes & yes & yes\\
\bottomrule
\end{tabular}
\end{table}
\begin{table}[!htbp]
\centering
\caption{Reward-matched mechanism ladder, five seeds. Aux-VFR is supervised but not fed to the action head; VFR-cond. adds that pathway.}
\label{tab:ladder}
\small
\begin{tabular}{lrrr}
\toprule
Variant & SR (\%) $\uparrow$ & Cov. viol. (\%) $\downarrow$ & VFR valid. (\%) $\uparrow$\\
\midrule
Backbone & 36.79$\pm$2.33 & 62.50$\pm$6.27 & --\\
Latent-5D & 45.89$\pm$8.14 & 25.56$\pm$17.55 & 17.60$\pm$10.98\\
Aux-VFR & 52.66$\pm$0.62 & 2.40$\pm$1.74 & 85.51$\pm$4.03\\
VFR-cond. & 54.65$\pm$4.35 & 2.73$\pm$4.40 & 79.32$\pm$3.52\\
w/o CGRA & 40.12$\pm$5.42 & 62.23$\pm$8.99 & 54.67$\pm$6.24\\
Strict bottleneck & 42.25$\pm$1.81 & 52.05$\pm$6.94 & 49.75$\pm$4.02\\
\textbf{Full} & \textbf{54.57$\pm$2.26} & \textbf{2.33$\pm$1.44} & 82.63$\pm$2.73\\
\bottomrule
\end{tabular}
\end{table}
\FloatBarrier
Full versus Backbone gives $+17.77$ success points ($p=5.5\times10^{-4}$) and $-60.17$ violation points ($p=4.1\times10^{-5}$). Aux-VFR versus Latent-5D gives $-23.16$ violation points ($p=0.042$); VFR-cond. versus Aux-VFR has no detected difference. Full versus w/o CGRA gives $+14.45$ success points ($p=0.0023$) and $-59.89$ violation points ($p<10^{-4}$). Separately, the coordinate-control study removes each named coordinate and tests random or permuted supervised targets. Success changes by only $0.57$--$1.74$ points and violation stays within $0.33$ points of Full, so no individual named coordinate is load-bearing in this benchmark. Figure~\ref{fig:app_ladder} visualizes the complete ladder.

\FloatBarrier

The coordinate controls in Table~\ref{tab:semantic_controls} make the scope of the semantic claim explicit:
\begin{table}[!htbp]
\centering
\caption{Five-seed semantic-coordinate controls. Small changes relative to Full rule out a claim that any single named coordinate is the primary performance driver.}
\label{tab:semantic_controls}
\scriptsize
\begin{tabular}{lrrr}
\toprule
Variant & SR (\%) & $\Delta$SR (pp) & Cov. viol. (\%)\\
\midrule
Full & 54.25$\pm$1.29 & 0.00 & 2.78$\pm$0.68\\
Drop coverage margin & 53.10$\pm$0.60 & -1.15 & 2.46$\pm$0.71\\
Drop energy & 53.21$\pm$0.94 & -1.04 & 2.97$\pm$0.70\\
Drop coverage-valid & 53.22$\pm$0.76 & -1.03 & 2.67$\pm$1.09\\
Drop compute & 53.22$\pm$0.37 & -1.03 & 2.54$\pm$0.43\\
Drop delay & 52.51$\pm$0.26 & -1.74 & 2.50$\pm$0.87\\
Random supervised target & 52.51$\pm$0.40 & -1.74 & 2.75$\pm$1.04\\
Permuted coordinates & 53.68$\pm$1.37 & -0.57 & 2.45$\pm$0.81\\
\bottomrule
\end{tabular}
\end{table}

\begin{figure}[!htbp]
\centering
\includegraphics[width=0.65\linewidth]{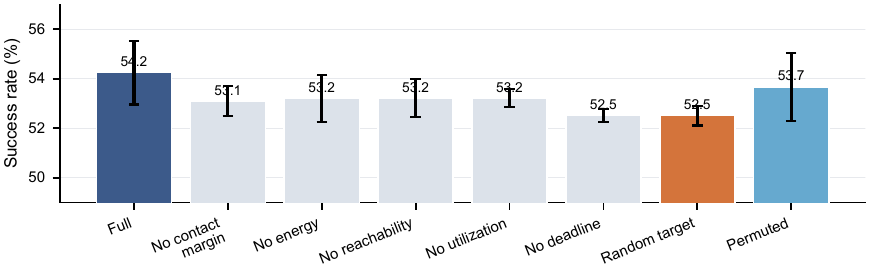}
\caption{Success under coordinate removal, random-target, and coordinate-permutation controls. Error bars are sample standard deviations over five seeds.}
\label{fig:semantic_controls}
\end{figure}

Figure~\ref{fig:semantic_controls} shows that no single coordinate removal creates a large success change.
\FloatBarrier

Table~\ref{tab:u4} crosses four representations with CGRA on/off over S10:
\begin{table}[!htbp]
\centering
\caption{S10 representation--CGRA factorial in SAGIN. COV is coverage violation.}
\label{tab:u4}
\scriptsize
\begin{tabular}{lrrrrr}
\toprule
Representation & CGRA SR & no-CGRA SR & $\Delta$SR & CGRA COV & no-CGRA COV\\
\midrule
Handcrafted & 53.36$\pm$0.79 & 38.45$\pm$3.08 & +14.91 & 2.37$\pm$0.63 & 62.51$\pm$5.27\\
Learned & 53.31$\pm$0.52 & 37.25$\pm$2.77 & +16.06 & 2.32$\pm$0.70 & 64.45$\pm$5.66\\
Random & 52.53$\pm$0.32 & 36.06$\pm$2.89 & +16.48 & 2.73$\pm$0.96 & 59.60$\pm$9.56\\
Latent-5D & 49.58$\pm$4.28 & 35.42$\pm$1.99 & +14.16 & 19.32$\pm$12.07 & 61.40$\pm$2.89\\
\bottomrule
\end{tabular}
\end{table}
\FloatBarrier
All four paired success effects exclude zero; the repeated-measures interaction test gives $p=0.323$. Random targets nearly match handcrafted targets when CGRA is active, while the latent head has substantially worse compliance. These observations support the two-part interpretation used in the paper: CGRA supplies most of the SAGIN performance gain, and grounding supplies the auditable feasibility interface.

\subsection{High-precision topology evaluation}
\label{app:topology}
The high-precision study runs 20 episodes for every seed--method--topology cell. Intervals in Fig.~\ref{fig:mechanism}b are seed-level 95\% confidence intervals; Table~\ref{tab:topology} reports mean$\pm$sample standard deviation.
\begin{table}[!htbp]
\centering
\caption{Absolute success rate (\%) over seven topologies.}
\label{tab:topology}
\scriptsize
\begin{tabular}{lrrr}
\toprule
Topology & VERA & AB-MAPPO-R & MAPPO-R\\
\midrule
Regional & 53.17$\pm$2.84 & 46.01$\pm$7.59 & 26.12$\pm$3.22\\
Regional deterministic & 53.05$\pm$0.21 & 50.00$\pm$3.42 & 23.03$\pm$3.07\\
Walker & 50.93$\pm$2.78 & 42.34$\pm$5.13 & 26.59$\pm$3.04\\
Walker-star & 51.42$\pm$2.94 & 40.68$\pm$6.08 & 26.29$\pm$2.90\\
Phase shift & 52.44$\pm$1.92 & 49.65$\pm$16.93 & 25.73$\pm$7.02\\
Inclination shift & 52.81$\pm$1.97 & 41.52$\pm$5.59 & 26.58$\pm$2.70\\
Sparse shell & 51.30$\pm$4.16 & 39.36$\pm$4.69 & 26.16$\pm$3.21\\
\bottomrule
\end{tabular}
\end{table}
\FloatBarrier
VERA--MAPPO-R differences are $24.33$--$30.02$ points and remain significant after Holm correction on all seven topologies. VERA--AB-MAPPO-R differences are $2.80$--$11.95$ points and directionally positive on all seven; none is Holm-significant. The latter is not described as a confirmed superiority claim.

\subsection{Cross-topology signed-margin calibration}
\label{app:calibration}
\begin{figure}[!htbp]
\centering
\includegraphics[width=0.55\linewidth]{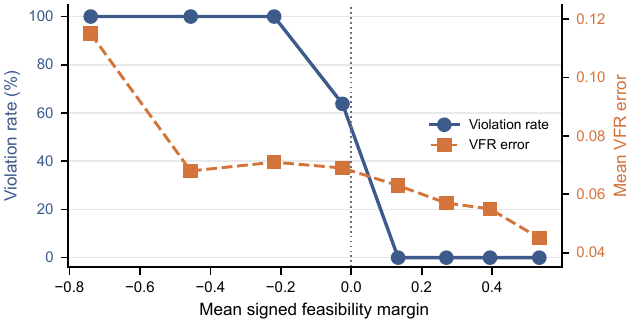}
\caption{Source-topology signed-margin calibration. The violation transition occurs around the feasibility boundary; the preceding table verifies the same ordering on four shifted topologies.}
\label{fig:app_margin_calibration}
\end{figure}

The signed-margin study evaluates 80,118 decisions on each of Walker, Walker-star, inclination-shift, and sparse-shell, then bins $m=\min(m_{coverage},m_{delay})$ into eight quantiles. For every topology, the three most negative bins have 100\% violation, the boundary bin has 52.4--86.7\%, and all four positive bins have 0\%. Table~\ref{tab:calibration} reports the aggregated bins.
\begin{table}[!htbp]
\centering
\caption{Signed-margin calibration aggregated over the source topology.}
\label{tab:calibration}
\scriptsize
\begin{tabular}{crrrr}
\toprule
Bin & Margin mean & VFR error & Coverage flip (\%) & Violation (\%)\\
\midrule
0 & -0.740 & 0.115 & 8.5 & 100.0\\
1 & -0.456 & 0.068 & 0.3 & 100.0\\
2 & -0.219 & 0.071 & 0.3 & 100.0\\
3 & -0.025 & 0.069 & 0.4 & 63.8\\
4 & 0.133 & 0.063 & 0.0 & 0.0\\
5 & 0.270 & 0.057 & 0.0 & 0.0\\
6 & 0.394 & 0.055 & 0.0 & 0.0\\
7 & 0.534 & 0.045 & 0.0 & 0.0\\
\bottomrule
\end{tabular}
 \end{table}
Violation is monotone non-increasing across bins in all four topologies. VFR error is largest in the most negative region and decreases overall toward positive margins, though it need not decrease at every adjacent bin. Figure~\ref{fig:app_margin_calibration} shows the source-topology transition.

\FloatBarrier

\subsection{VMAS transfer and normalized OOD audit}
\label{app:vmas}
The cross-domain study ports the training pattern to VMAS cooperative navigation. The five supervised quantities are collision margin, goal reachability, a safety indicator, speed margin, and time buffer. Table~\ref{tab:vmas_indomain} reports the in-domain result.
\begin{table}[!htbp]
\centering
\caption{VMAS in-domain performance over five training seeds.}
\label{tab:vmas_indomain}
\small
\begin{tabular}{lrr}
\toprule
Variant & Goal success (\%) & Training collision events (\%)\\
\midrule
Full & 24.75$\pm$4.09 & 0.64$\pm$0.92\\
Latent-5D & 28.41$\pm$3.69 & 0.26$\pm$0.37\\
No VFR / no CGRA & 4.86$\pm$4.20 & 0.51$\pm$0.59\\
Reward-only constrained & 1.56$\pm$2.96 & 0.73$\pm$1.02\\
\bottomrule
\end{tabular}
\end{table}
Full exceeds the no-VFR/no-CGRA backbone by $19.88$ points ($p=0.0020$), while Latent-5D exceeds Full by $3.66$ points ($p=0.038$). The result isolates transfer of the verification/credit pattern and shows that learned latent coordinates remain competitive outside SAGIN.

\begin{figure}[!htbp]
\centering
\includegraphics[width=0.85\linewidth]{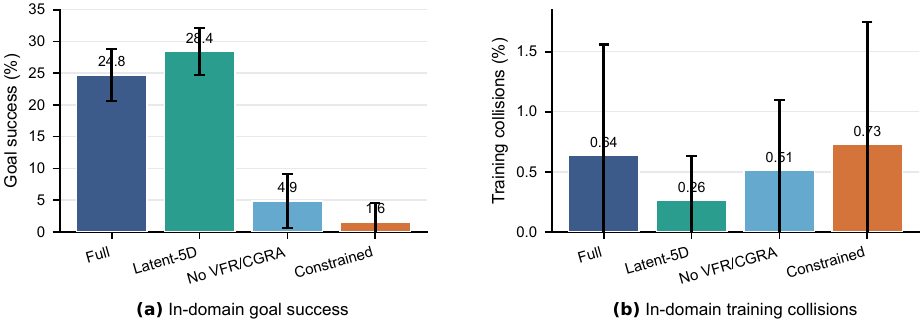}
\caption{VMAS in-domain performance over five training seeds: (a) goal success and (b) training collision events. In-domain collisions stay low for every variant, so the safety difference is driven by the out-of-distribution audit. The latent representation exceeds the hand-specified semantic head, which is retained as a negative result for universal semantic superiority.}
\label{fig:app_vmas_indomain}
\end{figure}

Figure~\ref{fig:app_vmas_indomain} visualizes the goal-success comparison and the in-domain collision rates.
\FloatBarrier

The separate OOD inference audit uses preserved checkpoints, the five new seeds, two shifts, and 100 episodes per cell. Its comma-separated values (CSV) file has 4,000 episode rows ($4$ variants $\times5$ seeds $\times2$ shifts $\times100$ episodes). Collision is $100\times$events/agent-steps, hence bounded and comparable across agent counts. Table~\ref{tab:vmas_ood} reports both shifts.
\begin{table}[!htbp]
\centering
\caption{VMAS OOD audit, mean$\pm$sample standard deviation across five inference seeds.}
\label{tab:vmas_ood}
\scriptsize
\begin{tabular}{lrrrr}
\toprule
Variant & Density goal & Density collision & Scale goal & Scale collision\\
\midrule
Full & 36.67$\pm$13.94 & 1.93$\pm$3.25 & 20.00$\pm$32.60 & 4.10$\pm$3.54\\
No VFR/CGRA & 20.00$\pm$21.73 & 2.60$\pm$4.56 & 10.00$\pm$13.69 & 0.00$\pm$0.00\\
Latent-5D & 53.33$\pm$21.73 & 6.60$\pm$11.20 & 50.00$\pm$39.53 & 7.50$\pm$6.55\\
Constrained & 23.33$\pm$22.36 & 7.80$\pm$7.15 & 20.00$\pm$27.39 & 1.60$\pm$3.05\\
\bottomrule
\end{tabular}
\end{table}
\FloatBarrier
Latent-5D has the highest goal rate on both shifts but also the highest collision rate; Full does not dominate every metric. Because this audit covers only the new five seeds and reuses trained checkpoints, it is kept separate from S10 training claims.

\subsection{Resource-allocation and force/contact failures}
\label{app:boundary}
The S10 resource-allocation factorial uses the same four representations and CGRA switch as the SAGIN factorial. Every CGRA arm reaches $63.47\%$ success and exactly $100\%$ capacity violation. Removing CGRA lowers success but restores compliance (Table~\ref{tab:resource_factorial}):
\begin{table}[!htbp]
\centering
\caption{S10 resource-allocation factorial.}
\label{tab:resource_factorial}
\scriptsize
\begin{tabular}{lrrrr}
\toprule
Representation & CGRA SR & no-CGRA SR & CGRA COV & no-CGRA COV\\
\midrule
Handcrafted & 63.47$\pm$0.06 & 49.73$\pm$1.25 & 100.00$\pm$0.00 & 5.70$\pm$11.59\\
Learned & 63.47$\pm$0.06 & 49.57$\pm$0.52 & 100.00$\pm$0.00 & 2.65$\pm$4.31\\
Random & 63.47$\pm$0.06 & 49.29$\pm$0.71 & 100.00$\pm$0.00 & 1.43$\pm$0.63\\
Latent-5D & 63.47$\pm$0.06 & 49.27$\pm$0.63 & 100.00$\pm$0.00 & 2.07$\pm$1.47\\
\bottomrule
\end{tabular}
\end{table}
\FloatBarrier
CGRA's success effects are $+13.74$--$+14.20$ points and its violation effects are $+94.30$--$+98.57$ points. Neither success nor violation shows a detected representation interaction ($p=0.564$ and $p=0.392$). This is reward hacking: proportional allocation and the flat over-capacity penalty make increased demand advantageous to the local counterfactual evaluator.

The force/contact study applies the pre-registered collapse rule---success below 5\% for five consecutive checkpoints after previously exceeding 20\%---to 50 S10 traces. Table~\ref{tab:force_collapse} gives the per-arm attribution.
\begin{table}[!htbp]
\centering
\caption{Force/contact collapse attribution over ten seeds per arm.}
\label{tab:force_collapse}
\scriptsize
\begin{tabular}{lrrr}
\toprule
Arm & Collapsed seeds & Converged SR (\%) & Converged COV (\%)\\
\midrule
No VFR/CGRA & 2/10 & 47.42$\pm$13.52 & 26.59\\
Full, no CGRA & 3/10 & 48.11$\pm$10.12 & 29.32\\
CGRA only & 10/10 & 2.44$\pm$1.65 & 4.63\\
Full & 10/10 & 0.87$\pm$1.68 & 4.07\\
Full, no annealing & 10/10 & 2.33$\pm$3.51 & 4.33\\
\bottomrule
\end{tabular}
\end{table}
All CGRA-bearing arms collapse, but five collapses also occur among 20 CGRA-absent traces. The attribution is therefore strong and heterogeneous, not an exclusive single-cause law.

\begin{figure}[!htbp]
\centering
\includegraphics[width=0.99\linewidth]{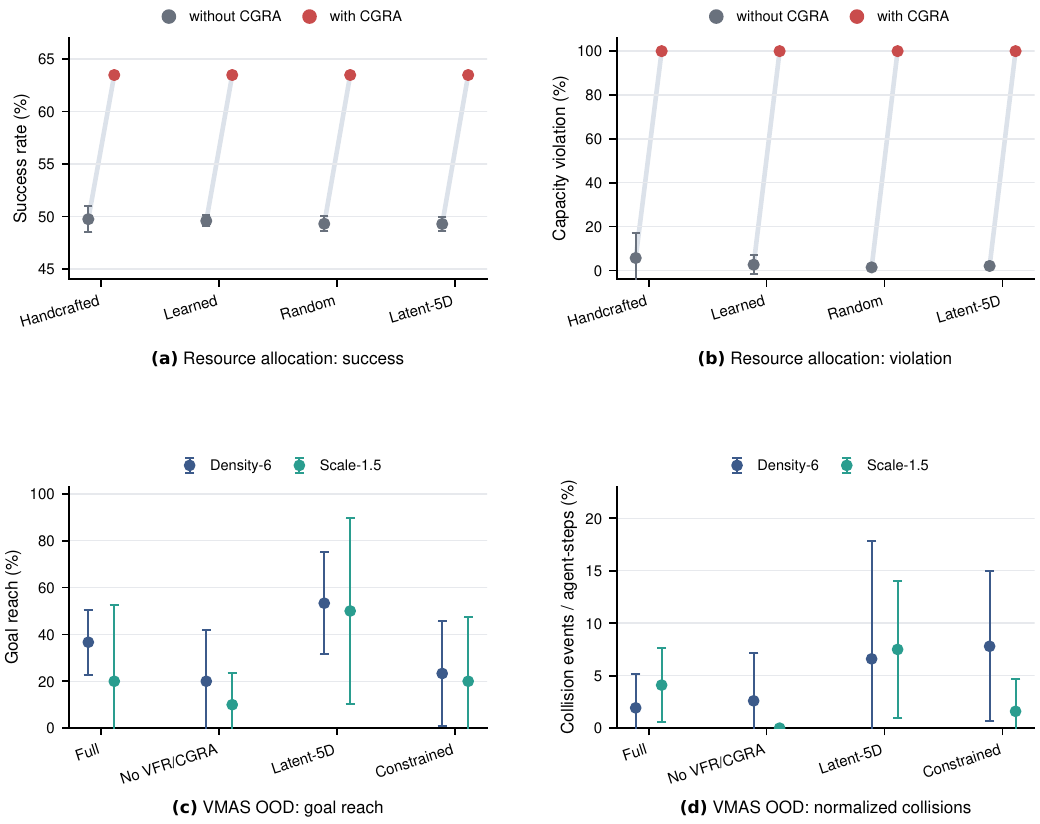}
\caption{\textbf{Measured transfer boundary.} S10 resource allocation gains success by violating shared capacity across all four representations (a,b). In the 4,000-row VMAS OOD audit, Latent-5D has the highest goal reach but also the highest normalized collision rate (c,d). Error bars are sample standard deviations across seeds.}
\label{fig:boundary}
\end{figure}

The stabilizer study tests standard-deviation flooring, a Huberized relative advantage, and externality correction under pre-registered gates. On SAGIN, all three pass on 10/10 seeds: success is $53.09$--$53.44\%$ and violation $2.05$--$2.54\%$. In resource allocation, all variants remain at $100\%$ violation and pass on 0/10 seeds. In force/contact, the best mean success is $4.06\%$ and every stabilizer passes on 0/10 seeds. Huberization is a no-op in both failed domains under the observed residuals; the externality term is also a no-op in force/contact. The unmodified CGRA is retained because none of the tested alternatives satisfies all gates. Figure~\ref{fig:boundary} combines the resource factorial with the VMAS OOD audit.

\FloatBarrier

\subsection{Inference diagnostics, scale, and deployment cost}
\label{app:diagnostics}
The inference pathway diagnostic uses three trained checkpoints and four held-out environment seeds, with 100 deterministic steps per arm. Table~\ref{tab:pathway} reports the interventions.
\begin{table}[!htbp]
\centering
\caption{Inference-pathway intervention results.}
\label{tab:pathway}
\small
\begin{tabular}{lrrr}
\toprule
Arm & SR (\%) & Cov. viol. (\%) & Target flip vs. Full (\%)\\
\midrule
Full & 61.98 & 0.73 & 0.0\\
Backbone-only & 62.43 & 0.34 & 3.5\\
VFR-only & 48.58 & 36.67 & 39.2\\
Both-zero & 35.56 & 40.33 & 45.6\\
\bottomrule
\end{tabular}
\end{table}
The actor was trained with both pathways; this is a diagnostic, not a retrained bottleneck comparison. Exact-autograd Jacobians remain locally sensitive to VFR, yet the backbone can preserve most decisions when explicit VFR input is zeroed. The ladder's retrained strict-bottleneck arm supplies the stronger intervention.

\begin{figure}[!htbp]
\centering
\includegraphics[width=0.90\linewidth]{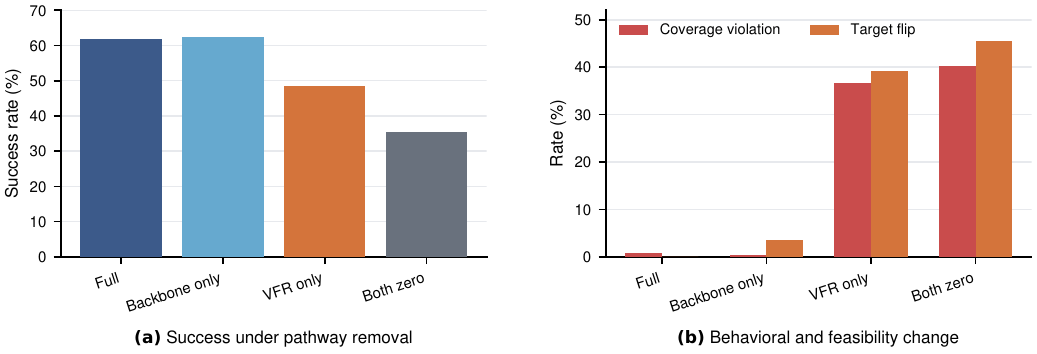}
\caption{Inference-time pathway decomposition. The table additionally reports the both-zero arm; this diagnostic is interpreted jointly with the retrained ladder.}
\label{fig:app_pathway}
\end{figure}

Figure~\ref{fig:app_pathway} visualizes the corresponding success and behavioral changes.
\FloatBarrier

Across user counts $K=10,20,40$ at six satellites, VERA success is 53.6/53.8/50.1\%, versus 37.6/34.9/26.8\% for AB-MAPPO and 34.0/32.1/35.7\% for MAPPO. Across satellite counts $M_{sat}=3,6,12$ at $K=20$, VERA success is 55.1/53.8/53.0\%; violation remains below 11\% across all six scales. Under zero-shot high load, link failure, a 0.10-s deadline, and heavier tasks, success retention is 100\%, 100\%, 47.8\%, and 46.3\%. The latter two shifts also increase the proportion of structurally infeasible tasks. Figure~\ref{fig:scale_deployment}a--b consolidates the scale and shift results.

On CPU, execution costs $0.939\pm0.106$ ms per step for VERA and $0.913\pm0.086$ ms for MAPPO; on CUDA, the values are $2.128\pm0.649$ and $2.073\pm0.668$ ms. A CGRA training update costs 6.94 ms on CPU and 8.81 ms on CUDA because it evaluates $G=8$ candidates. Candidate scoring and environment verification are absent from execution; Fig.~\ref{fig:scale_deployment}c shows the actor latency.

\subsection{Archive depth and experiment map}
\label{app:artifacts}
\enlargethispage{3\baselineskip}
The code/data archive separates immutable experiment results from manuscript-generation scripts. Table~\ref{tab:artifact_map} maps every headline result to its frozen summary and seed-level records.
\begin{table}[!htbp]
\centering
\caption{Experiment-to-artifact map in the companion code/data archive.}
\label{tab:artifact_map}
\scriptsize
\begin{tabular}{ll}
\toprule
Result & Archive path\\
\midrule
SAGIN representation factorial & \path{code/results/U4_identifiable_schema}\\
Topology evaluation & \path{data/U1_U6/U5_topology_20ep}\\
VMAS OOD audit & \path{data/E17_E24/E20_vmas_ood_normalized}\\
Force/contact attribution & \path{code_crossdomain/results/E24_force_control/_attrib5}\\
SAGIN stabilizers & \path{code/results/U6_stabilized_sagin}\\
Cross-domain stabilizers & \path{code_crossdomain/results/U6_stabilized_cgra}\\
Corrected resource factorial & \path{code_crossdomain/results_p1a_fixed}\\
\bottomrule
\end{tabular}
\end{table}

The learned-schema directory contains 15 per-seed CSV traces, a configuration snapshot, 15 checkpoints, 15 progress records, aggregate JavaScript Object Notation (JSON), and provenance metadata. All 48 manifest-listed files pass SHA-256 verification. That study does not include a learned-schema/no-CGRA arm; the resource factorial supplies the missing factorial cell in its own domain.